\documentclass[accepted]{article} 

\usepackage{times}

\usepackage{graphicx} 
\usepackage{subcaption}

\usepackage{natbib}

\usepackage{hyperref}       

\usepackage{booktabs}       
\usepackage{amsfonts,amssymb}       
\newcommand{\transpose}{\top}
\usepackage{nicefrac}       
\usepackage{microtype}      
\usepackage{multirow}

\usepackage{arydshln}

\usepackage{algorithm}
\usepackage[noend]{algorithmic}

\usepackage{xspace}
\usepackage{color}
\usepackage{amsmath}

\usepackage{enumitem}
\setitemize{noitemsep,topsep=0pt,parsep=0pt,partopsep=0pt,leftmargin=*}

\newcommand{\eg}{\hbox{\emph{e.g.}}\xspace}
\newcommand{\ie}{\hbox{\emph{i.e.}}\xspace}

\newcommand{\vs}{\hbox{\emph{vs.}}\xspace}

\usepackage{bm}
\usepackage[normalem]{ulem}
\usepackage[version=0.96]{pgf}
\usepackage{tikz}
\usetikzlibrary{arrows,shapes,snakes,automata,backgrounds,petri,positioning,shapes.geometric,decorations.pathreplacing}

\newcommand\hl[2]{%
    \tikz[baseline]%
    \definecolor{col}{RGB}{255,#1,#1}%
   \node[decorate,rectangle,fill=col,anchor=text,anchor=base west,%
   outer sep=1pt,inner xsep=.1pt,inner ysep=.8pt, rounded corners=2pt,%
   minimum height=\ht\strutbox+1pt,draw=black!50,line width=.4pt]{
		\strut{$#2$}\xspace
	};%
}%

\usepackage{listings}
\usepackage{icml2017}
\icmltitlerunning{Learning Continuous Semantic Representations of Symbolic Expressions}

\DeclareMathOperator{\children}{ch}
\DeclareMathOperator{\parent}{par}
\DeclareMathOperator*{\argmax}{arg\,max}

\begin{document}

\twocolumn[
\icmltitle{Learning Continuous Semantic Representations of Symbolic Expressions}



\icmlsetsymbol{equal}{*}

\begin{icmlauthorlist}
\icmlauthor{Miltiadis Allamanis}{msrc}
\icmlauthor{Pankajan Chanthirasegaran}{edi}
\icmlauthor{Pushmeet Kohli}{deepmind}
\icmlauthor{Charles Sutton}{edi,ati}
\end{icmlauthorlist}

\icmlaffiliation{edi}{University of Edinburgh, UK}
\icmlaffiliation{deepmind}{DeepMind, London, UK}
\icmlaffiliation{msrc}{Microsoft Research, Cambridge, UK}
\icmlaffiliation{ati}{The Alan Turing Institute, London, UK}

\icmlcorrespondingauthor{Miltiadis Allamanis}{t-mialla@microsoft.com}


\vskip 0.3in
]



\printAffiliationsAndNotice{Work started when M. Allamanis was at Edinburgh.
This work was done while P. Kohli was at Microsoft.}

\newcommand{\tfidf}{\textbf{\textsc{Tf-Idf}}\xspace}
\newcommand{\gru}{\textbf{\textsc{Gru}}\xspace}
\newcommand{\stackrnn}{\textbf{Stack-augmented \textsc{Rnn}}\xspace}

\newcommand{\singlernn}{\textbf{1-layer \recNN}\xspace}
\newcommand{\doublernn}{\textbf{2-layer \recNN}\xspace}

\newcommand{\semVec}{\textsc{SemVec}\xspace}
\newcommand{\semVecs}{\textsc{SemVec}s\xspace}

\newcommand{\nodetype}{\tau}
\newcommand{\xor}{\mathbin{\oplus}}
\newcommand{\representation}{\vect{r}}

\newcommand{\recNN}{\textsc{TreeNN}\xspace}
\newcommand{\recNNs}{\textsc{TreeNN}s\xspace}
\newcommand{\equivnetwork}{neural equivalence networks\xspace}
\newcommand{\eqnet}{\textsc{EqNet}\xspace}
\newcommand{\eqnets}{\textsc{EqNet}s\xspace}

\newcommand{\eqConstraint}{\textsc{SubexpAe}\xspace}

\newcommand{\knnscore}[1]{score_{#1}}

\newcommand{\vect}[1]{\mathbf{#1}}

\newcommand{\booleanDset}[0]{\textsc{Bool}\xspace}
\newcommand{\polyDset}[0]{\textsc{Poly}\xspace}
\newcommand{\seentest}{\textsc{SeenEqClass}\xspace}
\newcommand{\unseentest}[0]{\textsc{UnseenEqClass}\xspace}

\begin{abstract}
Combining abstract, symbolic reasoning with continuous neural reasoning is a grand challenge of representation learning. As a step in this direction, we propose a new architecture, called \emph{\equivnetwork}, for the problem of learning continuous semantic representations of algebraic and logical expressions. These networks are trained to represent semantic equivalence, even of expressions that are syntactically very different. The challenge is that  semantic representations must be computed in a syntax-directed manner,  because semantics is compositional, but at the same time, small changes in syntax can lead to very large changes in semantics, which can be difficult for continuous neural architectures. We perform an exhaustive evaluation  on the task of checking equivalence on a highly diverse class of symbolic algebraic and boolean expression types, showing that our model significantly outperforms existing architectures.

\end{abstract}

\section{Introduction}
Combining abstract, symbolic reasoning with continuous neural reasoning is a grand challenge of representation learning.
This is particularly important while dealing with exponentially large domains such as source code and logical expressions.
Symbolic notation allows us to abstractly represent a large set of states that may
be perceptually very different. 
Although symbolic reasoning is very powerful, it also tends to be hard. For example, problems such as
the satisfiablity of boolean expressions and automated formal proofs tend to be NP-hard or worse.
This raises the exciting opportunity of using pattern recognition within symbolic reasoning, that is, to learn patterns from 
datasets of symbolic expressions that approximately represent
semantic relationships.
However, apart from some notable exceptions \citep{alemi2016deepmath,loos2017deep,zaremba2014learning2},
this area has received relatively little attention in machine learning.
In this work, we explore the direction of 
learning continuous \emph{semantic} representations of symbolic expressions. The goal is for expressions with similar semantics to have similar continuous representations,
even if their syntactic representation is very different. Such representations have the potential to allow a new class of symbolic
reasoning methods based on heuristics that depend on the continuous representations, for example, by guiding a search procedure
in a symbolic solver based on a distance metric
in the continuous space. 
In this paper, we make a first essential step of addressing the problem of learning continuous semantic representations 
(\semVecs) for symbolic expressions. 
Our aim is, given access to a training set of pairs of expressions for which semantic equivalence is known,
to assign continuous vectors to symbolic expressions in such a way that semantically equivalent, but syntactically
diverse expressions are assigned to identical (or highly similar) continuous vectors. This is an important but hard
problem; learning composable \semVecs of symbolic expressions requires that we learn about the semantics of symbolic
elements and operators and how they map to the continuous representation space, thus encapsulating implicit
knowledge about symbolic semantics and its recursive abstractive nature.
As we show in our evaluation, relatively simple logical and polynomial expressions present significant
challenges and their semantics cannot be sufficiently represented by existing neural network architectures.

Our work in similar in spirit to the work of \citet{zaremba2014learning2}, who
focus on learning expression representations to aid the search for computationally efficient identities.
They use recursive neural networks (\recNN)\footnote{To avoid confusion, we use \recNN for \emph{recursive}
neural networks and RNN for \emph{recurrent} neural networks.} \citep{socher2012semantic} for modeling \emph{homogenous, single-variable} polynomial
expressions. While they present impressive results, we find that the \recNN model fails when applied to more
complex symbolic polynomial and boolean expressions. In particular, in our experiments we find that \recNNs tend
to assign similar representations to syntactically similar expressions, even when they are semantically very different.
The underlying conceptual problem is how to develop a continuous representation that follows syntax but \emph{not too much}, that respects
compositionality while also representing the fact that a small syntactic change can be a large semantic one.

To tackle this problem, we propose a new architecture,
called \emph{\equivnetwork} (\eqnet).
\eqnets learn how syntactic composition recursively composes \semVecs, like a \recNN, but are also designed
to model large changes in semantics as the network progresses up the syntax tree.
As equivalence is transitive, we formulate an objective function for training
based on equivalence classes rather than pairwise decisions. The network architecture is based on composing 
residual-like multi-layer networks, which allows more flexibility in modeling the semantic mapping up the syntax tree. To encourage  representations within an equivalence
class to be tightly clustered, we also introduce a training method that we call \emph{subexpression autoencoding},
which uses an autoencoder to force the representation of each subexpression to be predictable and reversible from its
syntactic neighbors.
Experimental evaluation on a highly diverse class of symbolic algebraic and boolean expression
types shows that \eqnets dramatically outperform existing architectures like \recNNs and RNNs.

To summarize, the main contributions of our work are:
(a) We formulate the problem of learning continuous semantic representations (\semVecs) of symbolic expressions and develop benchmarks for this task.
(b) We present \equivnetwork (\eqnets), a neural network architecture that learns
    to represent expression semantics onto a continuous semantic representation
    space and how to perform symbolic operations in this space.
(c) We provide an extensive evaluation
   on boolean and polynomial expressions, showing that \eqnets perform
   dramatically better than state-of-the-art alternatives.
Code and data are available at \href{http://groups.inf.ed.ac.uk/cup/semvec}{\texttt{groups.inf.ed.ac.uk/cup/semvec}}.

\section{Model}
In this work, we are interested in learning semantic, compositional representations
of mathematical expressions, which we call \semVecs, and in learning to generate identical representations
for expressions that are \emph{semantically} equivalent, \ie they belong to the
same equivalence class. Equivalence is a stronger property than similarity, which
has been the focus of previous work in neural network learning \cite{chopra2005learning}, since equivalence is additionally a
transitive relationship.

\paragraph{Problem Hardness.} Finding the equivalence of
arbitrary symbolic expressions is a NP-hard problem or worse. For example, if we focus on
boolean expressions, reducing an expression to the representation of
the \texttt{false} equivalence class amounts to proving its non-satisfiability
--- an NP-complete problem. Of course, we do \emph{not} expect
to circumvent an NP-complete problem with neural networks. A network for solving boolean equivalence
would require an exponential number of nodes in the size of the expression if $P \neq NP$.
Instead, our goal is to develop architectures that efficiently learn to solve the equivalence
problems for expressions that are similar to a smaller number of expressions in a given 
training set. The supplementary material
shows a sample of such expressions that illustrate the hardness of this
problem.

\paragraph{Notation and Framework.} To allow our 
representations to be compositional, we employ the general framework of
recursive neural networks (\recNN) \citep{socher2012semantic,socher2013recursive},
in our case operating on
tree structures of the syntactic parse of a formula. Given a tree $T$, \recNNs learn distributed
representations for each node in the tree by recursively combining the representations of its subtrees
using a neural network.
We denote the children of a node $n$ as $\children(n)$ which is a (possibly empty)
ordered tuple of nodes. We also use $\parent(n)$ to refer to the
parent node of $n$. Each node in our tree has a type, \eg a terminal node could
be of type ``\texttt{a}'' referring to the variable $a$ or of type ``\texttt{and}'' referring to
a node of the logical \textsc{And} ($\land$) operation. We refer to the type of a node $n$ as $\nodetype_n$.
\newcommand{\rnnNetFunc}{\textsc{TreeNN}\xspace}%
\newcommand{\rnnCombine}{\textsc{Combine}\xspace}%
In pseudocode, \recNNs retrieve the
representation of a tree $T$ rooted at node $\rho$,
by invoking the function $\rnnNetFunc(\rho)$ that returns a vector representation
$\representation_\rho \in \mathbb{R}^D$, \ie, a \semVec. The function is defined as

\rnnNetFunc(current node $n$)
\begin{algorithmic}
\IF{$n$ \textbf{is not} a leaf}
	  \STATE $\representation_n \gets \rnnCombine(\rnnNetFunc(c_0), \dots, \rnnNetFunc(c_k), \nodetype_n)$, where $\left(c_0, \dots, c_k \right)=\children(n)$
  \ELSE
    \STATE $\representation_n \gets \textsc{LookupLeafEmbedding}(\nodetype_n)$
  \ENDIF
  \STATE \textbf{return} $\representation_n$
\end{algorithmic}\vspace{-.5em}
The general framework of \rnnNetFunc allows two points of variation, the
implementation of \textsc{LookupLeafEmbedding} and \rnnCombine. 
Traditional \recNNs \citep{socher2013recursive} define
\textsc{LookupLeafEmbedding} as a simple lookup operation within a matrix
of embeddings and \rnnCombine as a single-layer neural network.
As discussed next, these will both prove to be serious limitations in our setting.
To train these networks to learn \semVecs,
we will use a supervised objective based on a set of known equivalence relations
(see \autoref{subsec:training}).
\subsection{Neural Equivalence Networks}

\newcommand{\layer}{\bar{l}}
\begin{figure*}[t]\centering
	\begin{subfigure}[b]{\textwidth}\centering
		\includegraphics[width=.9\textwidth]{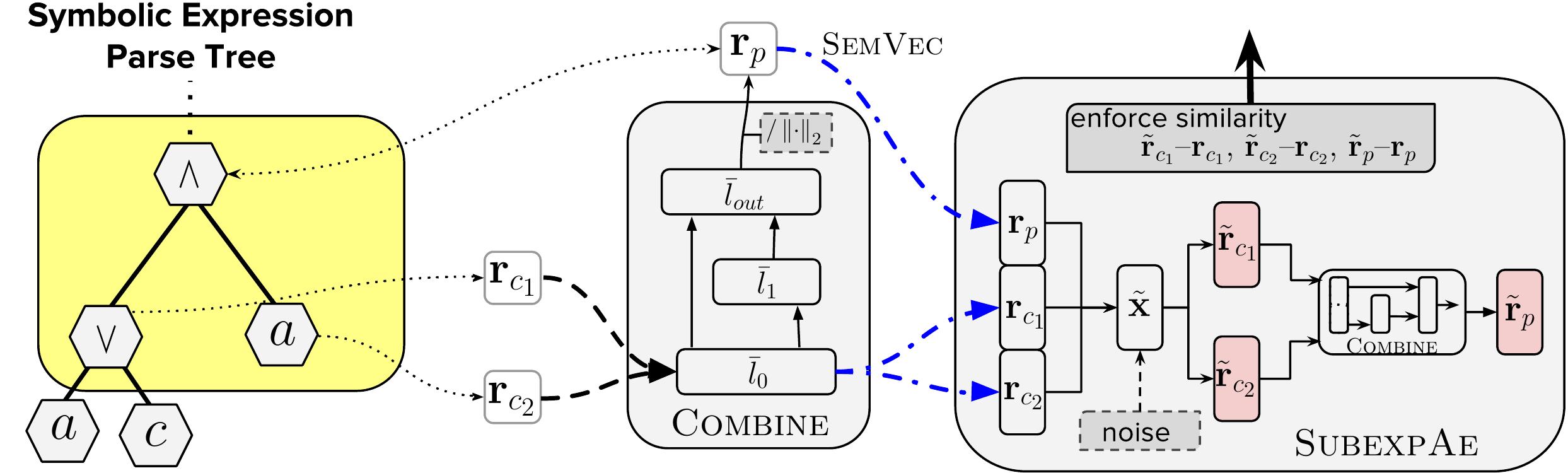}
		\caption{Architectural diagram of \eqnets. Example parse tree shown is of the boolean expression $(a\lor c)\land a.$}\label{fig:arch}
	\end{subfigure}\vspace{.5em}

	\begin{subfigure}[b]{.4\textwidth}
		\rnnCombine($\representation_{c_0}, \dots, \representation_{c_k}, \nodetype_p$)

		\begin{algorithmic}
		  \STATE $ \layer_0 \gets [\representation_{c_0}, \dots, \representation_{c_k}]$
		  \STATE $\layer_1 \gets \sigma \left(W_{i,\nodetype_p} \cdot \layer_0 \right)$
		  \STATE $\layer_{out} \gets W_{o0,\nodetype_p} \cdot \layer_0 + W_{o1,\nodetype_p} \cdot \layer_1$
		  \STATE \textbf{return} $\layer_{out} / \left\lVert\layer_{out}\right\rVert_2$
		\end{algorithmic}
		\caption{\rnnCombine of \eqnet.}\label{fig:eqnet:combine}
	\end{subfigure}
	\begin{subfigure}[b]{.59\textwidth}
		\eqConstraint($\representation_{c_0}, \dots, \representation_{c_k}, \representation_p, \nodetype_p$)

		\begin{algorithmic}
			\STATE $\vect{x} \gets [\representation_{c_0}, \dots, \representation_{c_k}]$
			\STATE $\vect{\tilde{x}} \gets \tanh \left(W_d \cdot \tanh \left( W_{e, \nodetype_p} \cdot [\representation_p, \vect{x}] \cdot \vect{n} \right)\right)$
			\STATE $\vect{\tilde{x}} \gets \vect{\tilde{x}} \cdot \left\lVert\vect{x}\right\rVert_2 / \left\lVert\vect{\tilde{x}}\right\rVert_2$
			\STATE $\tilde{\representation}_p \gets \rnnCombine(\vect{\tilde{x}}, \nodetype_p)$
			\STATE \textbf{return} $-\left(\vect{\tilde{x}}^\transpose \vect{x} + \tilde{\representation}^\transpose_p\representation_p\right)$
		\end{algorithmic}
		\caption{Loss function used for subexpression autoencoder}\label{fig:eqnet:eqConstraint}
	\end{subfigure}\vspace{-1em}
	\caption{\eqnet architecture.}\vspace{-1.5em}
\end{figure*}

Our domain requires that the network learns to abstract away syntax, assigning 
identical representations to expressions
that may be syntactically different but semantically equivalent, and also
assigning different representations to expressions that may be syntactically very similar but 
non-equivalent. In this work, we find that standard neural architectures do
not handle well this challenge. To represent semantics from
syntax, we need to learn to recursively compose and decompose semantic representations
and remove syntactic ``noise''. Any syntactic operation may significantly change
semantics (\eg negation, or appending $\land \textsc{False}$) while we may reach the
same semantic state through many possible operations. This necessitates using
high-curvature operations over the semantic representation space. Furthermore, some
operations are semantically reversible
and thus we need to learn reversible semantic representations (\eg $\lnot\lnot A$ and $A$
should have an identical \semVecs).
Based on these, we define \emph{\equivnetwork} (\eqnet), which learn to
compose representations of equivalence classes into new equivalence classes (\autoref{fig:arch}).
Our network follows the \recNN architecture, \ie is implemented
using \rnnNetFunc to model
the compositional nature of symbolic expressions but is adapted based on the
domain requirements.
The extensions we introduce have two aims: first, to improve
the network training; and second, and more interestingly,
to encourage the learned representations to abstract away surface level
information while retaining semantic content.

The first extension that we introduce is to the network structure
at each layer in the tree. Traditional
\recNNs \citep{socher2013recursive} use a single-layer
neural network at each tree node. During our preliminary investigations and
in \autoref{sec:evaluation}, we found that single layer networks are \emph{not}
adequately expressive to capture all operations that transform the input
\semVecs to the output \semVec and maintain semantic equivalences, requiring high-curvature operations. Part
of the problem stems from the fact that within the Euclidean space of
\semVecs some operations need to be non-linear. For example
a simple \textsc{xor} boolean operator requires high-curvature operations
in the continuous semantic representation space.
 Instead, we turn to
multi-layer neural networks. In particular, we define the network
as shown in the function
\rnnCombine in \autoref{fig:eqnet:combine}. This uses
a two-layer MLP with a residual-like connection to compute the \semVec of each parent node in that syntax
tree given that of its children. Each node type $\nodetype_n$, \eg, each logical operator,
has a different set of weights. We  experimented
with deeper networks but this did not yield any improvements.

However, as \recNNs become deeper, they suffer from
optimization issues, such as diminishing and exploding gradients. This is essentially because
of the highly compositional nature of tree structures, where the same network
(\ie the \rnnCombine non-linear function) is used recursively, causing it to ``echo''
its own errors and producing unstable feedback loops.
We observe this problem even with only two-layer MLPs, as the overall
network can become quite deep when using two layers for each node
in the syntax tree.
We resolve this issue in the training procedure by
constraining each \semVec to have unit norm. That is, we set
$\textsc{LookupLeafEmbedding}(\nodetype_n) = C_{\nodetype_n} / \left\lVert C_{\nodetype_n} \right\rVert_2,$
and we normalize the output of the final layer of \rnnCombine in \autoref{fig:eqnet:combine}.
The normalization step
of $\layer_{out}$ and $C_{\nodetype_n}$ is somewhat similar to weight normalization \citep{salimans2016weight}
and vaguely resembles layer normalization \citep{ba2016layer}.
Normalizing the \semVecs
partially resolves issues with diminishing and exploding gradients,
and removes a spurious degree of freedom in the semantic representation.
As simple as this modification may seem, we found it vital for obtaining
good performance, and all of our multi-layer \recNNs converged to
low-performing settings without it.

Although these modifications seem to improve the representation
capacity of the network and its ability to be trained, we found
that they were not on their own sufficient for good performance. 
In our early experiments, we
noticed that the networks were primarily focusing on syntax instead of semantics, \ie, expressions that were nearby in the continuous
space were primarily ones that were syntactically similar.
At the same time, we observed that the networks did not learn to unify representations
of the same equivalence class, observing multiple syntactically distinct but semantically
equivalent expressions to have distant \semVecs.

Therefore we modify the training objective in order to encourage
the representations to become more abstract, reducing their
dependence on surface-level syntactic information. 
We add a regularization term on the \semVecs that we call a 
\emph{subexpression autoencoder} (\eqConstraint).
We design this regularization to encourage
the \semVecs to have two properties: abstraction
and reversibility.
Because \emph{abstraction} arguably means removing irrelevant
information, a network with a bottleneck layer
seems natural, but we want the training objective to encourage
the bottleneck to discard syntactic information rather than
semantic information. To achieve this, we introduce a component
that aims to encourage \emph{reversibility}, which we explain by an example.
Observe that given the semantic representation
of any two of the three nodes of a subexpression (by 
which we mean the parent, left child, right child of an expression
tree) it is often possible to completely determine or 
at least
place strong constraints on the semantics of the third. For example, 
consider a boolean formula $F(a,b) = F_1(a,b) \lor F_2(a,b)$ where $F_1$ and $F_2$ are arbitrary propositional formulae over the variables $a,b$. Then clearly if we know that $F$ implies that $a$ is true but $F_1$ does not, then $F_2$ must imply that $a$ is true.
More generally, if $F$  belongs to some equivalence class $e_0$
and $F_1$ belongs to a different class $e_1$, we want the continuous
representation of $F_2$ to reflect that there are strong constraints
on the equivalence class of $F_2$. 
 
Subexpression autoencoding
encourages abstraction by
employing an autoencoder with a bottleneck, thereby removing
irrelevant information from the representations, 
and encourages reversibility by autoencoding the
 parent and child representations together, to encourage dependence in the representations of parents and children.
More specifically, given any node $p$ in the tree with children $c_0 \ldots c_k$,
we can define a parent-children
tuple $[\representation_{c_0}, \dots, \representation_{c_k}, \representation_p]$
containing the (computed) \semVecs of the children and parent nodes.
What \eqConstraint does is to autoencode this representation tuple
into a low-dimensional space with a denoising
autoencoder.  We then seek to minimize the reconstruction error of the child
representations ($\tilde{\representation}_{c_0},\dots,\tilde{\representation}_{c_k}$)
as well as the reconstructed parent representation $\tilde{\representation}_p$
that can be computed from the reconstructed children. More formally, we minimize
the return value of \eqConstraint in \autoref{fig:eqnet:eqConstraint}
where $\vect{n}$ is a binary noise vector with $\kappa$ percent of its elements
set to zero. Note that the encoder is specific to the
parent node type $\nodetype_p.$ Although our \eqConstraint may seem similar to the
recursive autoencoders of \citet{socher2011semi}, it differs in two major ways.
First, \eqConstraint autoencodes on the entire parent-children
representation tuple, rather than the child representations alone. Second,  the encoding is \emph{not} 
used to compute the parent representation, but only serves as a regularizer.

Subexpression autoencoding has several desirable effects. First, it forces each
parent-children tuple to lie in a low-dimensional space, requiring the network to compress
information from the individual subexpressions. Second, because the denoising autoencoder is reconstructing
parent and child representations together, this encourages child representations to be predictable
from parents and siblings. Putting these two together, the goal is that the information discarded by the autoencoder
bottleneck will be more syntactic than semantic, assuming that the semantics of child node is more predictable
from its parent and sibling than its syntactic realization. The goal is to nudge the network to learn consistent,
reversible semantics. Additionally, subexpression autoencoding has the potential to gradually unify distant representations that
belong to the same equivalence class. 
To illustrate this point, imagine
two semantically equivalent $c_0'$ and $c_0''$ child nodes of different expressions that have
distant \semVecs, \ie $\left\lVert\representation_{c_0'} - \representation_{c_0''}\right\rVert_2\gg\epsilon$
although $\rnnCombine(\representation_{c_0'}, \dots) \approx \rnnCombine(\representation_{c_0''}, \dots).$
In some cases due to the autoencoder noise, the differences between the
input tuple $\vect{x'}, \vect{x''}$ that contain $\representation_{c_0'}$ and $\representation_{c_0''}$
will be non-existent and the decoder will predict a single location $\tilde{\representation}_{c_0}$
(possibly different from $\representation_{c_0'}$ and $\representation_{c_0''}$).
Then, when minimizing the reconstruction error, both $\representation_{c_0'}$ and $\representation_{c_0''}$
will be attracted to $\tilde{\representation}_{c_0}$ and eventually should merge.

\subsection{Training}
\label{subsec:training}
We train \eqnets from a dataset of expressions whose semantic equivalence
is known. Given a training set $\mathcal{T} = \{ T_1 \ldots T_N \}$ of parse trees of expressions,
we assume that the training set is partitioned into equivalence classes
$\mathcal{E} = \{ e_1 \ldots e_J \}$.
We use a supervised objective similar to classification; the difference between classification and our setting
is that whereas standard classification problems consider a fixed set of class labels, in our setting the number
of equivalence classes in the training set will vary with $N$.
Given an
expression tree $T$ that belongs to the equivalence class $e_i \in \mathcal{E}$, we
compute the probability
\begin{align}
  P(e_i|T) = \frac{\exp\left(\recNN(T)^\transpose\vect{q}_{e_i} + b_i\right)}{\sum_j \exp\left(\recNN(T)^\transpose\vect{q}_{e_j}+b_j\right)}
\end{align}
where $\vect{q}_{e_i}$ are model parameters that we can interpret as
representations of each equivalence class that appears in the training class,
and $b_i$ are scalar bias terms.
Note that in this work, we only use information about the equivalence class
of the whole expression $T$, ignoring available information about subexpressions.
This is without loss of generality, because if we do know the equivalence class
of a subexpression of $T$, we can simply add that subexpression to the training set.
To train the model, we use a max-margin objective that maximizes classification accuracy, \ie
\begin{align}
	\mathcal{L}_\textsc{Acc}(T, e_i) =  \max\left(0, \argmax_{e_j \neq e_i, e_j \in \mathcal{E}} \log \frac{P(e_j|T)}{P(e_i|T)} + m\right)
\end{align}
where $m>0$ is a scalar margin.
And therefore the optimized loss function for a single expression tree $T$ that belongs to
equivalence class $e_i\in\mathcal{E}$ is
\begin{align}
  \mathcal{L}(T, e_i) = \mathcal{L}_\textsc{Acc}(T, e_i) +\frac{\mu}{|Q|}\sum_{n \in Q}\eqConstraint(\children(n), n)
\end{align}
where $Q=\left\{n \in T: |\children(n)|>0\right\}$, \ie contains the non-leaf nodes of $T$ and $\mu\in(0,1]$ a
scalar weight.
We found that subexpression autoencoding is counterproductive
early in training, before the \semVecs begin to represent aspects of semantics.
So, for each epoch $t$, we set $\mu=1-10^{-\nu t}$ with $\nu \geq 0$.
Instead of the supervised objective that we propose,
an alternative option for training \eqnet would be a Siamese
objective \citep{chopra2005learning} that learns about similarities (rather than
equivalence) between expressions. In practice, we found the optimization
to be very unstable, yielding suboptimal performance. We believe that this has
to do with the compositional and recursive nature of the task that creates
unstable dynamics and the fact that equivalence is a stronger property than
similarity.

\section{Evaluation}
\label{sec:evaluation}
\newcommand{\simpleBooleanOps}[0]{$\land$, $\lor$, $\lnot$\xspace}
\newcommand{\allBooleanOps}[0]{$\land$, $\lor$, $\lnot$, $\xor$, $\Rightarrow$\xspace}
\newcommand{\simplePolyOps}[0]{$+$, $-$\xspace}
\newcommand{\polyOps}[0]{$+$, $-$, $\cdot$\xspace}
\begin{table*}
  \caption{Dataset statistics and results. \textsc{Simp} datasets
  contain simple operators (``\simpleBooleanOps'' for \booleanDset
  and ``\simplePolyOps'' for \polyDset) while the rest contain all operators (\ie
  ``\allBooleanOps'' for \booleanDset and ``\polyOps'' for \polyDset). $\xor$ is the
  \textsc{Xor} operator. The number in the dataset name indicates its expressions' maximum
  tree size. \textsc{L} refers
  to a ``larger'' number of 10 variables.
  $H$ is the entropy of equivalence classes.}\label{tbl:dataset}\footnotesize\centering
  \begin{tabular}{lrrrrrrrrrr} \toprule
    Dataset &  \#  & \# Equiv & \# & $H$ & \multicolumn{6}{c}{$\knnscore{5}$ (\%) in \unseentest}\\
          &   Vars   & Classes & Exprs & &tf-idf & GRU & StackRNN & 1L \recNN & 2L \recNN & \eqnet \\\midrule
    \textsc{SimpBool8} & 3 & 120 & 39,048 & 5.6 & 17.4 & 30.9 & 26.7 & 27.4 & 25.5 & \textbf{97.4}\\
    \textsc{SimpBool10}$^S$ & 3 & 191 & 26,304 & 7.2 & 6.2 & 11.0 & 7.6 & 25.0 &93.4 & \textbf{99.1}\\
    \textsc{Bool5} & 3 & 95 & 1,239 & 5.6 & 34.9 & 35.8 & 12.4 & 16.4 & 26.0 & \textbf{65.8}\\
    \textsc{Bool8} & 3 & 232 & 257,784 & 6.2 & 10.7 & 17.2 & 16.0 & 15.7 & 15.4 & \textbf{58.1}\\
    \textsc{Bool10}$^S$ & 10 & 256 & 51,299 & 8.0 & 5.0 & 5.1 & 3.9 & 10.8 & 20.2 & \textbf{71.4}\\
    \textsc{SimpBoolL5} & 10 & 1,342 & 10,050& 9.9 & 53.1 & 40.2 & 50.5 & 3.48 & 19.9 & \textbf{85.0}\\
    \textsc{BoolL5} & 10 & 7,312 & 36,050 & 11.8 & 31.1 & 20.7 & 11.5 & 0.1 & 0.5 & \textbf{75.2} \\
    \textsc{SimpPoly5} & 3 & 47 & 237 & 5.0 & 21.9 & 6.3 & 1.0 & 40.6 & 27.1 & \textbf{65.6}\\
    \textsc{SimpPoly8} & 3 & 104 & 3,477 & 5.8 & 36.1 & 14.6 & 5.8 & 12.5 & 13.1 & \textbf{98.9}\\
    \textsc{SimpPoly10} & 3 & 195 & 57,909 & 6.3 & 25.9 & 11.0 & 6.6 & 19.9 & 7.1 & \textbf{99.3}\\
    \textsc{oneV-Poly10} & 1 & 83 & 1,291 & 5.4 & 43.5 & 10.9 & 5.3 & 10.9 & 8.5 & \textbf{81.3}\\
    \textsc{oneV-Poly13} & 1 & 677 & 107,725 & 7.1 & 3.2 & 4.7 & 2.2 & 10.0 & 56.2 & \textbf{90.4}\\
    \textsc{Poly5} & 3 & 150 & 516 & 6.7 & 37.8 & 34.1 & 2.2 & 46.8 & \textbf{59.1} & 55.3\\
    \textsc{Poly8} & 3 & 1,102 & 11,451 & 9.0 & 13.9 & 5.7 & 2.4 & 10.4 & 14.8 & \textbf{86.2}\\
  \bottomrule\end{tabular}

  $^S$~dataset contains all equivalence classes but at most 200 uniformly sampled (without replacement) expressions per equivalence class.\vspace{-1em}
\end{table*}

\paragraph{Datasets.} We generate datasets
of expressions grouped into equivalence classes from two domains. The datasets from the \booleanDset domain
contain boolean expressions and the \polyDset datasets contain polynomial expressions.
In both domains, an expression is either a variable, a binary operator that combines
two expressions, or a unary operator applied to a single expression. When defining equivalence, we interpret distinct variables as referring
to different entities in the domain, so that, e.g., the polynomials $c \cdot (a \cdot a + b)$
and   $f \cdot (d \cdot d + e)$ are not equivalent.
For each domain, we generate ``simple'' datasets which use a smaller set
of possible operators and ``standard'' datasets which use a larger set of more complex operators.
We generate each dataset by exhaustively generating \emph{all} parse trees
up to a maximum tree size. All expressions are symbolically simplified into a canonical from
in order to determine their equivalence class and
are grouped accordingly. \autoref{tbl:dataset} shows the datasets we generated.
In the supplementary material we present some sample expressions.
For the polynomial domain, we also generated
\textsc{OneV-Poly} datasets, which are polynomials over a single variable, since they are similar to the setting considered by \citet{zaremba2014learning2}
--- although \textsc{OneV-Poly} is still a little more general because it is not
restricted to homogeneous polynomials.
Learning \semVecs for boolean expressions is already a hard problem; with $n$ boolean variables,
there are $2^{2^n}$ equivalence classes (\ie one for each possible truth table).
We split the datasets into training, validation and test sets. We create two
test sets, one to measure generalization performance on equivalence classes that were
seen in the training data (\seentest), and one to measure generalization to unseen equivalence classes (\unseentest).
It is easiest to describe \unseentest first.
To create the \unseentest, we randomly select 20\% of all the equivalence classes,
and place all of their expressions in the test set. We select equivalence classes only if they contain
at least two expressions but less than three times the average number of
expressions per equivalence class. We thus
avoid selecting very common (and hence trivial to learn) equivalence classes in the testset.
Then, to create \seentest, we take the remaining 80\% of the equivalence classes,
and randomly split the expressions in each class into training, validation, \seentest test
in the proportions 60\%--15\%--25\%. We provide the datasets online at \href{http://groups.inf.ed.ac.uk/cup/semvec}{\texttt{groups.inf.ed.ac.uk/cup/semvec}}.

\paragraph{Baselines.}
To compare the performance of our model, we train the following baselines.
\tfidf: learns a representation given the expression tokens (variables,
operators and parentheses). This captures topical/declarative knowledge but is unable to
capture procedural knowledge. \gru refers to the token-level gated recurrent unit
encoder of \citet{bahdanau2014neural} that encodes the token-sequence of an
expression into a distributed representation.
\stackrnn refers to the work of \citet{joulin2015inferring} which was used to
learn algorithmic patterns and uses a stack as a memory and operates on the
expression tokens.
We also include two recursive neural networks (\recNN). The
\singlernn which is the original \recNN also used by \citet{zaremba2014learning2}.
We also include a \doublernn, where \textsc{Combine} is a classic two-layer MLP
without residual connections.
This shows the effect of \semVec normalization and subexpression autoencoder.

\paragraph{Hyperparameters.} We tune the hyperparameters of all models
using Bayesian optimization \citep{snoek2012practical}
on a boolean dataset with 5 variables and
maximum tree size of 7 (not shown in \autoref{tbl:dataset}) using the average $k$-NN
($k=1,\dots,15$) statistics (described next). The selected hyperparameters
are detailed in the supplementary material.

\subsection{Quantitative Evaluation}

\paragraph{Metrics.}
To evaluate the quality of the learned representations we count the proportion
of $k$ nearest neighbors of each expression (using cosine similarity) that
belong to the same equivalence class. More formally, given a test query
expression $q$ in an equivalence class $c$
we find the $k$ nearest neighbors $\mathbb{N}_k(q)$ of $q$ across all expressions,
and define the score as
\begin{align}
  \knnscore{k}(q) = \frac{\left|\mathbb{N}_k(q) \cap c\right|}{\min(k, |c|)}.
\end{align}
To report results for a given testset, we simply average $\knnscore{k}(q)$
for all expressions $q$ in the testset. We also report the precision-recall
curves for the problem of clustering the \semVecs into their appropriate equivalence 
classes.

\begin{figure*}[t]
  \begin{subfigure}[b]{.43\textwidth}\centering
    \includegraphics[width=\textwidth]{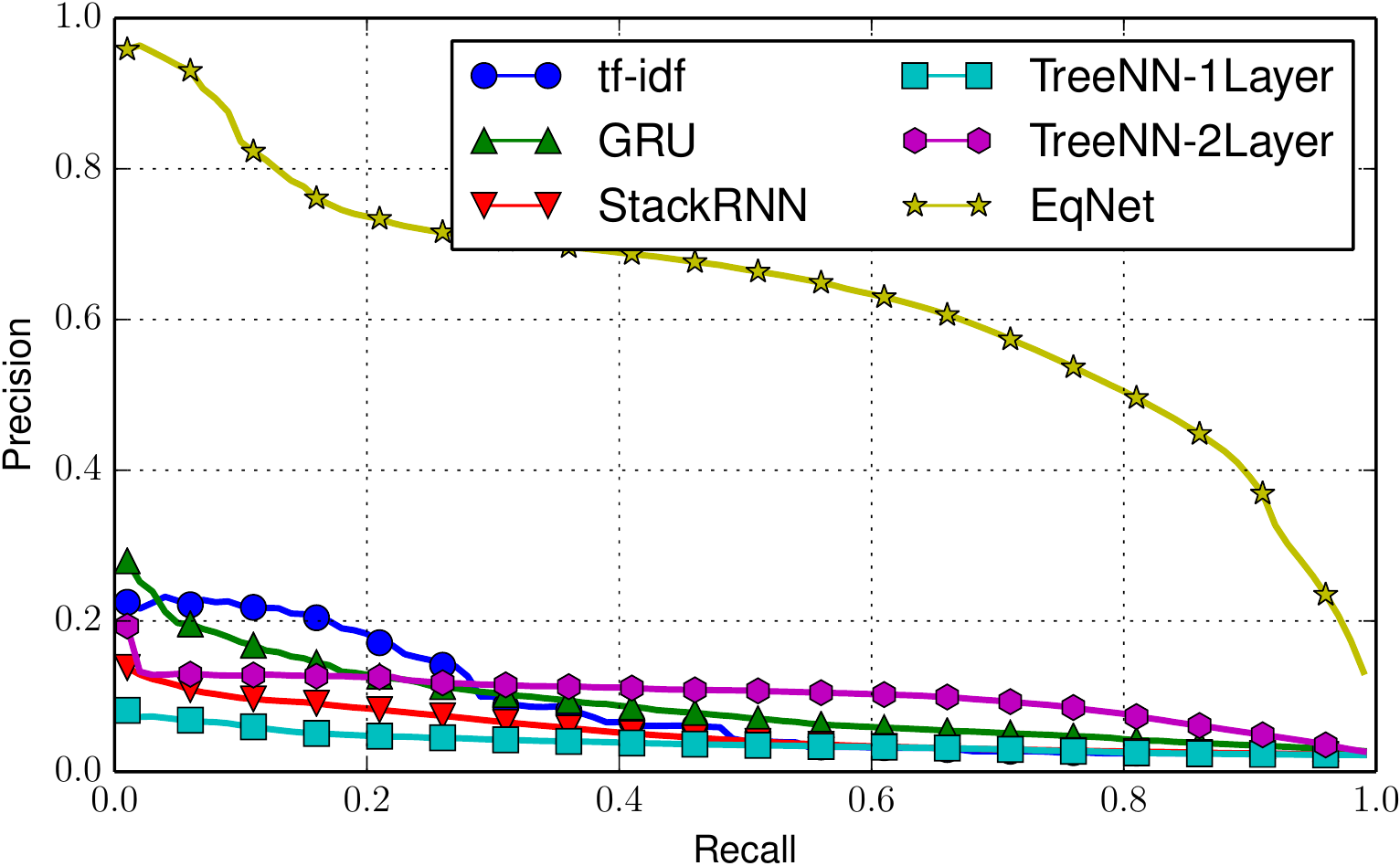}
    \caption{\seentest}\label{fig:prSeen}
  \end{subfigure}\hfill
  \begin{subfigure}[b]{.43\textwidth}\centering
    \includegraphics[width=\textwidth]{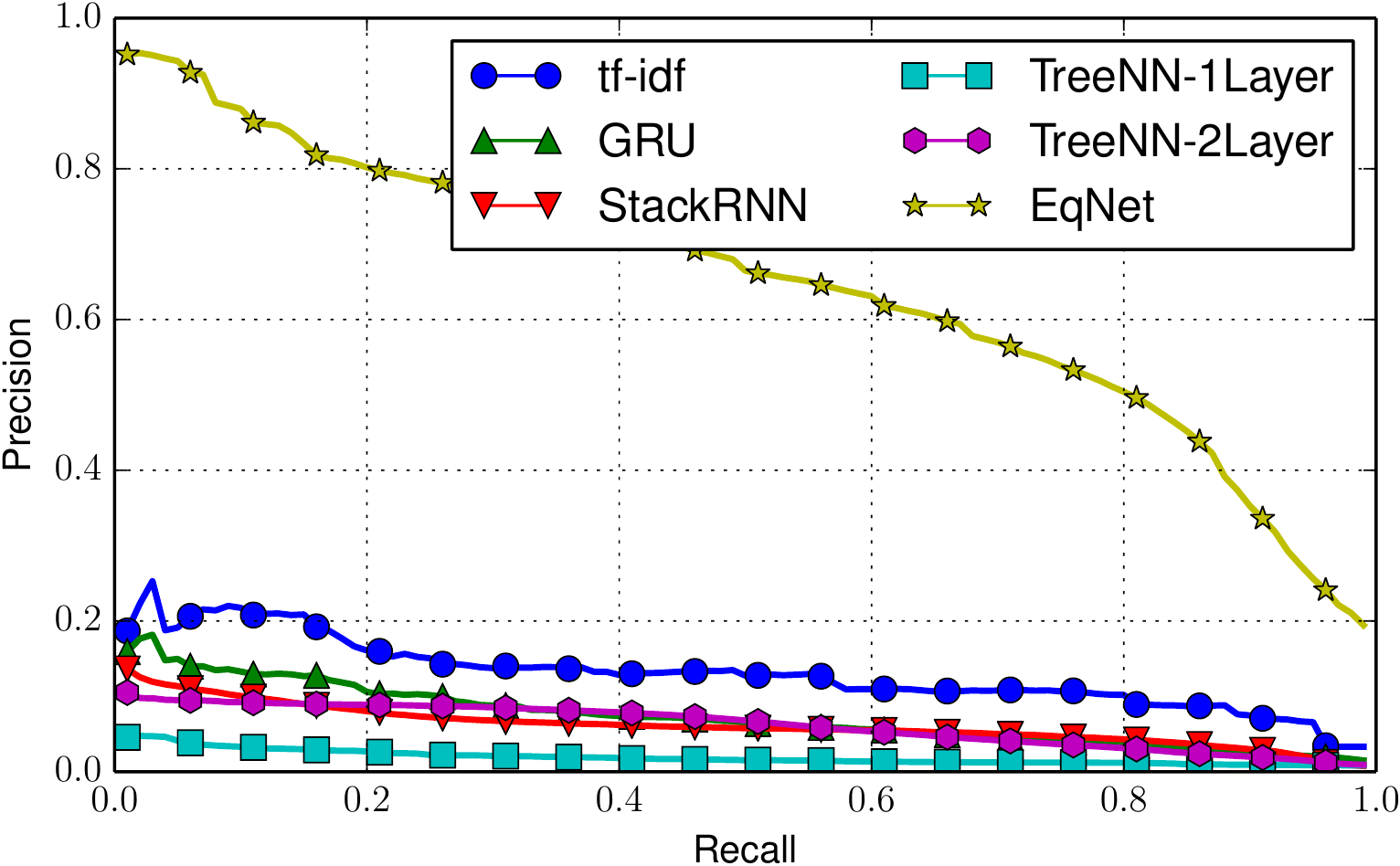}
    \caption{\unseentest}\label{fig:prUnseen}
  \end{subfigure}
  \caption{Precision-Recall Curves averaged across datasets.}\label{fig:prCurves}\vspace{-1em}
\end{figure*}

\begin{figure}[tb]\centering
    \begin{subfigure}[b]{.48\columnwidth}
      \includegraphics[width=.8\textwidth]{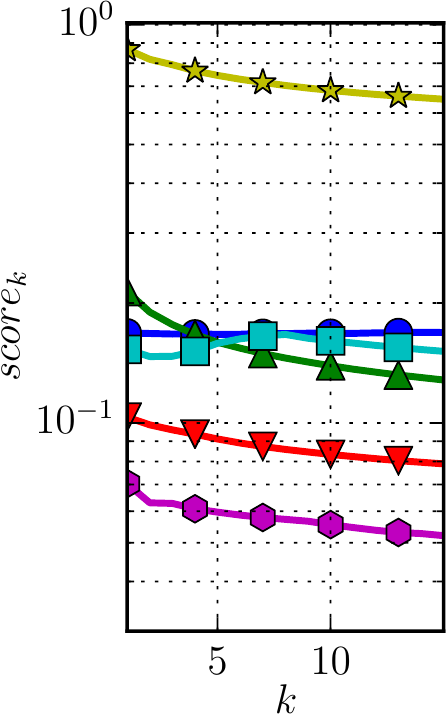}
      \caption{\seentest}\label{fig:knnEval:transfer:test}
    \end{subfigure}
    \begin{subfigure}[b]{.48\columnwidth}
      \includegraphics[width=.8\textwidth]{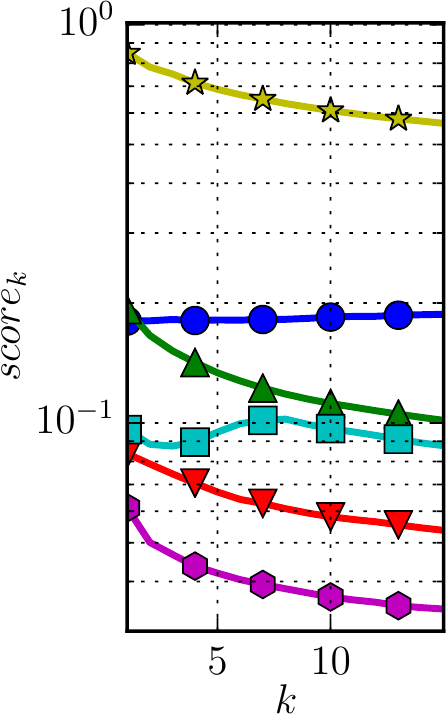}
      \caption{\unseentest}\label{fig:knnEval:transfer:neweq}
    \end{subfigure}
  \includegraphics[width=\columnwidth]{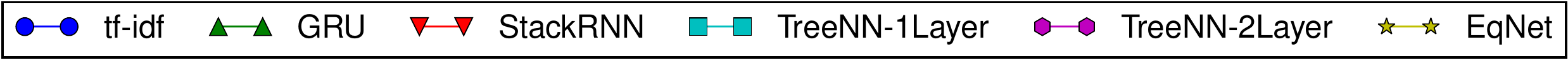}
  \caption{Evaluation of compositionality; training set simpler than test set. 
  Average $\knnscore{k}$ ($y$-axis in log-scale).
  Markers are shown every three ticks for clarity. \recNN
  refers to \citet{socher2012semantic}.}\label{fig:knnEval}\vspace{-1em}
\end{figure}

\paragraph{Evaluation.}
\autoref{fig:prCurves} presents the average per-model precision-recall curves across
the datasets. \autoref{tbl:dataset} shows $\knnscore{5}$ of \unseentest.
Detailed plots are found in the supplementary material.
\eqnet performs better for all
datasets, by a large margin. The only exception is \textsc{Poly5}, where the 2-L \recNN performs better.
However, this may have to do with the small size of the dataset. The reader
may observe that the simple datasets (containing fewer operations and variables) are easier
to learn. Understandably, introducing more variables increases the size of the
represented space reducing performance. The tf-idf method performs better in settings
with more variables, because it captures well the variables and
operations used. Similar observations can be made for sequence models.
The one and two layer \recNNs have mixed performance; we believe that this has
to do with exploding and diminishing gradients due to the deep and highly
compositional nature of \recNNs.
Although \citet{zaremba2014learning2} consider a different problem to us,
they use data similar to the \textsc{OneV-Poly} datasets with a traditional \recNN
architecture. Our evaluation suggests that \eqnets perform much better within the
\textsc{OneV-Poly} setting.

\paragraph{Evaluation of Compositionality.} We evaluate whether \eqnets
successfully learn to compute compositional representations, rather than overfitting
to expression trees of a small size. To do this we consider a type of transfer setting,
in which we train on simpler datasets, but test on more complex
ones; for example, training on the training set of \textsc{Bool5}
but testing on the testset of \textsc{Bool8}. We average
over 11 different train-test pairs (full list in supplementary material)
and show the results in
\autoref{fig:knnEval:transfer:test} and \autoref{fig:knnEval:transfer:neweq}.
These graphs again show that \eqnets are better than any of the other
methods, and indeed, performance is only a bit worse than in the non-transfer setting.

\paragraph{Impact of \eqnet Components}
\eqnets differ from traditional \recNNs in two major ways, which we analyze
here. First, \eqConstraint improves performance. When training
the network with and without \eqConstraint, on average,
the area under the curve (AUC) of
$\knnscore{k}$ decreases by 16.8\% on the \seentest and 19.7\% on the \unseentest.
This difference is smaller in the transfer setting, where AUC decreases by 8.8\% on average.
However, even in this setting we observe that \eqConstraint
helps more in large and diverse datasets.
The second key difference to traditional \recNNs is the output normalization and the residual connections. Comparing
our model to the one-layer and two-layer \recNNs again, we find that output normalization results
in important improvements (the two-layer \recNNs have on average 60.9\% smaller
AUC). We note that only the combination of the residual connections and the output normalization
improve the performance, whereas when used separately, there are no significant improvements
over the two-layer \recNNs.

\subsection{Qualitative Evaluation}

\begin{figure}\centering
  \input{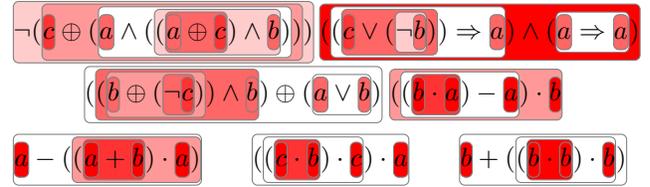}
  \caption{Visualization of $\knnscore{5}$ for all expression nodes for three \textsc{Bool10} and four \textsc{Poly8}
  test sample expressions using \eqnet. The darker
  the color, the lower the score, \ie white implies a score of 1 and dark
  red a score of 0.}\label{lbl:treeKnnViz}\vspace{-1.5em}
\end{figure}

\begin{table*}[t]
  \centering\caption{\emph{Non} semantically equivalent first nearest-neighbors
   from \textsc{Bool8} and \textsc{Poly8}. A checkmark indicates that the method correctly results in the nearest
    neighbor being from the same equivalence class.}\label{tbl:confusedExamples}\scriptsize
   \begin{tabular}{lcccccc}\toprule
    Expr & $a\land(a\land(a\land(\lnot c)))$ & $a\land(a\land(c\Rightarrow(\lnot c)))$ & $(a\land a)\land (c \Rightarrow (\lnot c))$& $a+(c\cdot (a+c))$ & $((a+c)\cdot c)+a$ & $(b\cdot b)-b$ \\ \midrule
    tfidf & $c\land((a\land a)\land (\lnot a))$ & $c\Rightarrow(\lnot((c\land a)\land a))$ & $c\Rightarrow(\lnot((c\land a)\land a))$& $a+(c+a)\cdot c$ & $(c\cdot a)+(a+c)$ & $b\cdot (b-b)$\\
    \textsc{Gru} & \checkmark & $a\land(a\land(c\land (\lnot c)))$ & $(a\land a)\land(c\Rightarrow (\lnot c))$ & $b+(c\cdot (a+c))$ & $((b+c)\cdot c)+a$ & $(b+b)\cdot b -b$\\
    1L-\recNN & $a\land(a\land(a\land(\lnot b)))$ & $a\land(a\land(c\Rightarrow(\lnot b)))$ & $(a\land a)\land(c\Rightarrow (\lnot b))$ & $a+(c\cdot (b+c))$ & $((b+c)\cdot c)+a$ & $(a-c)\cdot b - b$\\
    \eqnet & \checkmark & \checkmark & $(\lnot(b\Rightarrow (b\lor c)))\land a$ & \checkmark & \checkmark & $(b\cdot b)\cdot b - b$\\ \bottomrule
   \end{tabular}\vspace{-1em}
\end{table*}

\autoref{tbl:confusedExamples} shows
expressions whose \semVec nearest neighbor is of an expression of another equivalence class.
Manually inspecting boolean expressions, we find that \eqnet confusions happen
more when a \textsc{Xor} or implication operator is involved.
In fact, we fail to find any confused expressions for \eqnet \emph{not} involving these
operations in \textsc{Bool5} and in the top 100 expressions in \textsc{Bool10}.
As expected, tf-idf confuses expressions with others that contain the same
operators and variables ignoring order. In
contrast, \textsc{Gru} and \recNN tend to confuse expressions with very similar
symbolic representations, \ie that differ in one or two deeply nested variables or
operators. In contrast, \eqnet tends to confuse fewer expressions (as we
previously showed) and the confused expressions tend to be more syntactically
diverse and semantically related.

\autoref{lbl:treeKnnViz} shows a visualization of $\knnscore{5}$ for each
node in the expression tree. One may see that as \eqnet knows how to compose expressions
that achieve good score, even if the subexpressions achieve a worse score.
This suggests that for common
expressions, (\eg single variables and monomials) the network tends to
select a unique location, without merging the equivalence classes or
affecting the upstream performance of the network.
Larger scale interactive t-SNE visualizations can
be found at \href{http://groups.inf.ed.ac.uk/cup/semvec}{online}.

\begin{figure*}[t]\centering
  \begin{subfigure}[b]{.99\columnwidth}
    \includegraphics[width=\textwidth]{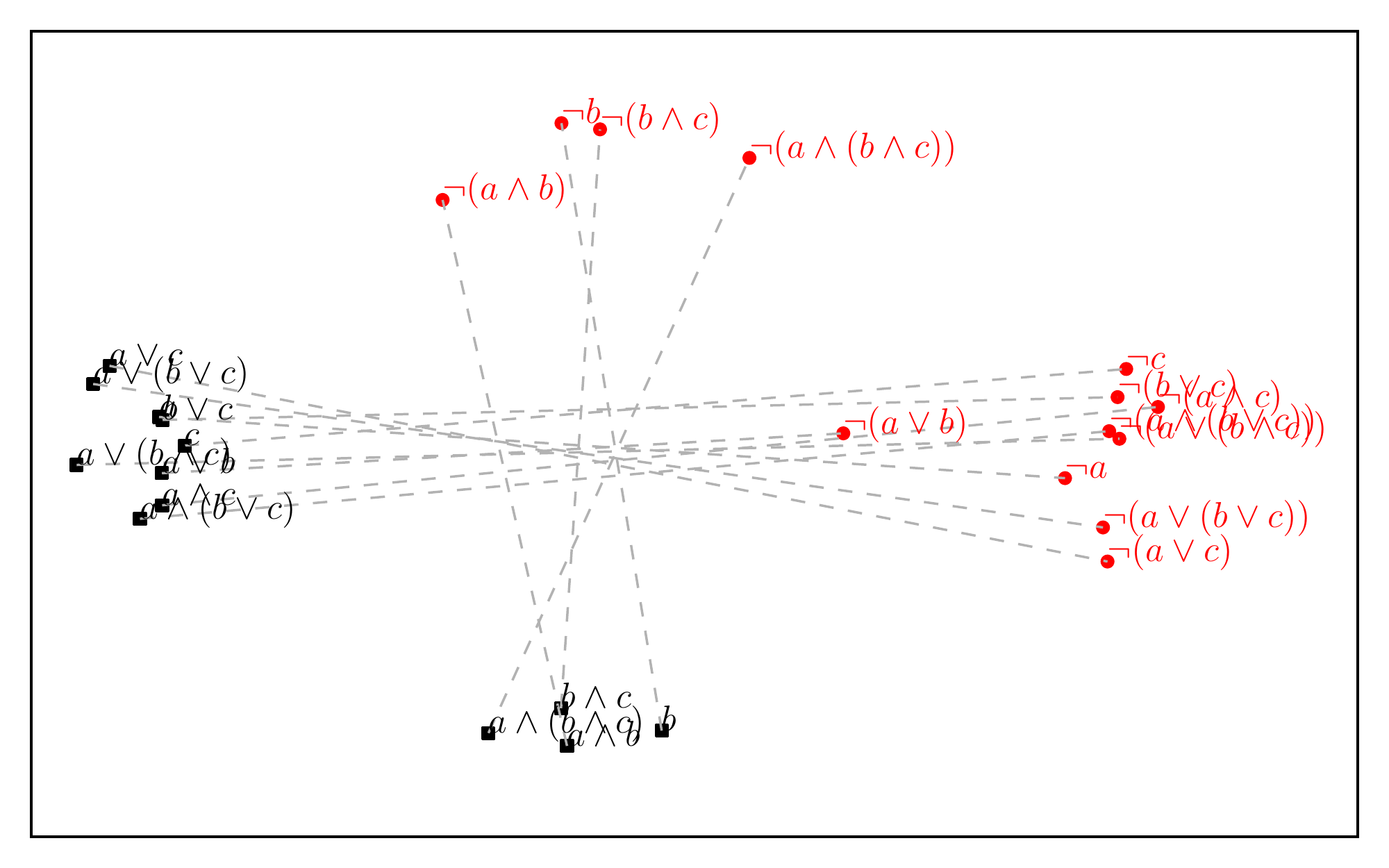}
    \caption{Negation in \booleanDset expressions}\label{fig:negationVecs:boolean}
  \end{subfigure}
  \begin{subfigure}[b]{0.99\columnwidth}
    \includegraphics[width=\textwidth]{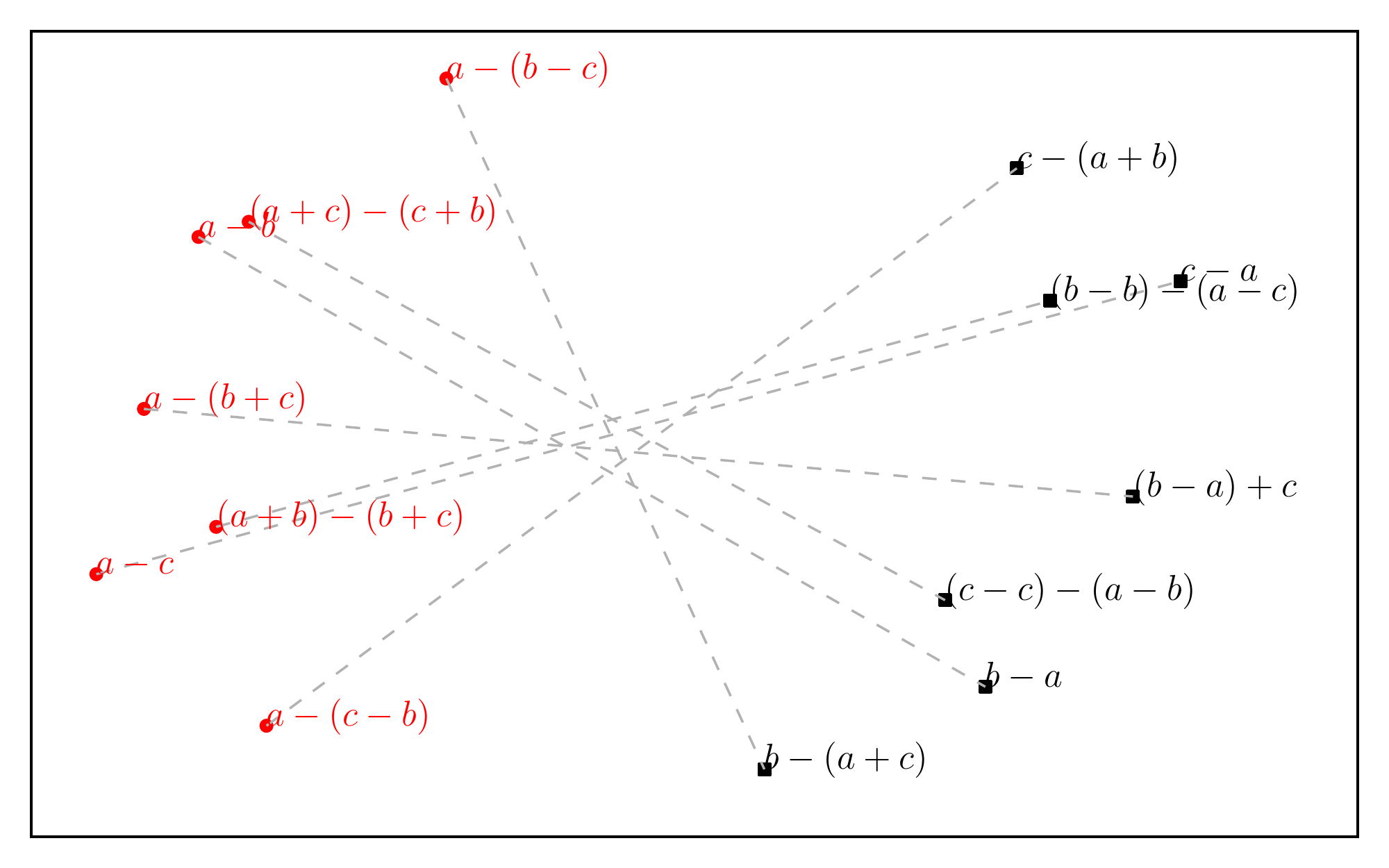}
    \caption{Negatives in \polyDset expressions}\label{fig:negationVecs:poly}
  \end{subfigure}
  \caption{A PCA visualization of some simple \emph{non}-equivalent boolean and polynomial expressions (black-square)
  and their negations (red-circle). The lines connect the negated expressions.}\label{fig:negationVecs}\vspace{-1em}
\end{figure*}

\autoref{fig:negationVecs} presents two PCA visualizations of the \semVecs
of simple expressions and their negations/negatives. It
can be discerned that the black dots and their negations (in red) are
discriminated in the semantic representation space. \autoref{fig:negationVecs:poly}
shows this property in a clear manner: left-right discriminates between polynomials
with $1$ and $-a$, top-bottom between polynomials with $-b$ and $b$
and the diagonal parellelt to $y=-x$ between $c$ and $-c$. We observe a similar
behavior in \autoref{fig:negationVecs:boolean} for boolean expressions.

\section{Related Work}
Researchers have proposed compilation schemes that can transform any given
program or expression to an equivalent neural network \citep{gruau1995neural,NETO2003,Siegelmann94}.
One can consider a serialized version of the resulting neural network as a representation of the expression. However, it is
not clear how we could compare the serialized representations corresponding to two expressions and whether this mapping
preserves semantic distances.

Recursive neural networks (\recNN) \citep{socher2012semantic,socher2013recursive}
have been successfully used in NLP with multiple applications. \citet{socher2012semantic}
show that \recNNs can learn to compute the \emph{values} of some simple propositional
statements. \eqnet's \eqConstraint may resemble recursive autoencoders \citep{socher2011semi}
but differs in form and function, encoding the whole parent-children tuple to
force a clustering behavior. In addition, when encoding each
expression our architecture does \emph{not} use a pooling layer but directly produces
a single representation for the expression.

\citet{mou2016convolutional} design tree convolutional networks
to classify code into student submission tasks. Although
they learn representations of the student tasks, these
representations capture task-specific syntactic features rather than code
semantics.  \citet{piech2015learning} also
learn distributed matrix representations of student code submissions.
However, to learn the representations, they use input and
output program states and do not test for program equivalence. Additionally,
these representations do not necessarily represent
program equivalence, since they do not learn the representations over
all possible input-outputs. \citet{allamanis2016convolutional}
learn variable-sized representations of source code snippets to summarize them with
a short function-like name but aim learn
summarization features in code rather than representations of
symbolic expression equivalence.

More closely related is the work of \citet{zaremba2014learning2} who
use a \recNN to guide the search for more efficient
mathematical identities, limited to homogeneous single-variable polynomial expressions.
In contrast, \eqnets consider at a much wider set of expressions, employ
subexpression autoencoding to guide the learned \semVecs to better represent equivalence,
and do \emph{not} use search when
looking for equivalent expressions.
\citet{alemi2016deepmath} use RNNs and convolutional neural networks to
detect features within mathematical expressions to speed the search for premise
selection in automated theorem proving but do not explicitly
account for semantic equivalence. In the future, \semVecs may be
useful within this area.

Our work is also related to recent work on neural network architectures
that learn controllers/programs \citep{gruau1995neural,graves2014neural,joulin2015inferring,grefenstette2015learning,dyer2015transition,reed2015neural,neelakantan2015neural,kaiser2015neural}.
In contrast to this work, we do not aim to learn how to evaluate expressions or execute programs
with neural network architectures but to learn continuous semantic representations (\semVecs) of
expression semantics irrespectively of how they are syntactically expressed
or evaluated.

\section{Discussion \& Conclusions}
In this work, we presented \eqnets, a first step in learning
continuous semantic representations (\semVecs) of procedural knowledge.
\semVecs have the potential of bridging continuous representations with
symbolic representations, useful in multiple applications in artificial intelligence,
machine learning and programming languages.

We show that \eqnets perform significantly better than state-of-the-art alternatives.
But further
improvements are needed, especially for more robust training of compositional models.
In addition, even for relatively small symbolic expressions, we have an
exponential explosion of the semantic space to be represented. Fixed-sized
\semVecs, like the ones used in \eqnet, eventually limit the capacity that is available to
represent procedural knowledge. In the future, to represent more complex procedures,
variable-sized representations would seem to be required.

\section*{Acknowledgments}
This work was supported by Microsoft Research through
its PhD Scholarship Programme and the Engineering and Physical Sciences
Research Council [grant number EP/K024043/1]. We thank the University of
Edinburgh Data Science EPSRC Centre for Doctoral Training for providing
additional computational resources.

\bibliographystyle{icml2017}
\bibliography{bibliography}

\appendix

\section{Synthetic Expression Datasets}
\label{appendix:datasetsamples}
\autoref{tbl:bool8data} and \autoref{tbl:poly8data} are sample expressions within an equivalence class for the two
types of datasets we consider.

\begin{table*}[tb] \centering
\begin{tabular}{ccc} \multicolumn{3}{c}{\textsc{Bool8}}\\ \toprule
    	$(\lnot a) \land (\lnot b)$ &  $(\lnot a \land \lnot c)\lor(\lnot b \land a \land c) \lor (\lnot c \land b)$ & $(\lnot a) \land b \land c$\\ \midrule
$a\lnot((\lnot a) \Rightarrow ((\lnot a) \land b))$ & $c \xor (((\lnot a) \Rightarrow a) \Rightarrow b)$ & $\lnot((\lnot b) \lor (( \lnot c) \lor a))$\\
	$\lnot((b \lor (\lnot (\lnot a))) \lor b)$ & $\lnot ((b \xor (b \lor a)) \xor c)$ &  $((a \lor b) \land c) \land (\lnot a)$\\
	$(\lnot a) \xor ((a \lor b) \xor a)$ & $\lnot ((\lnot (b \lor (\lnot a))) \xor c)$ & $(\lnot ((\lnot (\lnot b)) \Rightarrow a)) \land c$\\
  $(b \Rightarrow (b \Rightarrow a)) \land ( \lnot a)$ &  $((b \lor a) \xor (\lnot b)) \xor c)$ & $(c \land (c \Rightarrow (\lnot a))) \land b$\\
  $((\lnot a) \Rightarrow b) \Rightarrow (a \xor a)$ & $(\lnot ((b \xor a) \land a)) \xor c$ & $b \land (\lnot (b \land (c \Rightarrow a)))$\\
  \toprule
    	False & $(\lnot a) \land (\lnot b) \lor ( \land c)$ & $\lnot a \lor b$\\ \midrule
	$(a \xor a) \land (c \Rightarrow c)$ & $(a \Rightarrow (\lnot c)) \xor ( a \lor b)$ & $a \Rightarrow ((b \land (\lnot c)) \lor b)$\\
	$(\lnot b) \land (\lnot (b \Rightarrow a))$ & $(a \Rightarrow (c \xor b)) \xor b$ & $\lnot (\lnot (( b \lor a) \Rightarrow b))$ \\
	$b \land ((a \lor a) \xor a)$ & $b \xor (a \Rightarrow (b \xor c))$ & $(\lnot a) \xor (\lnot (b \Rightarrow ( \lnot a)))$\\
  $((\lnot b) \land b) \xor (a \xor a)$ & $(b \lor a) \xor (x \Rightarrow (\lnot a))$ & $b \lor (\lnot ((\lnot b) \land a))$\\
  $c \land (( \lnot (a \Rightarrow a)) \land c)$ & $b \xor ((\lnot a) \lor (c \xor b))$ & $\lnot ((a \Rightarrow (a \xor b)) \land a)$ \\
  \bottomrule\end{tabular} 
  \caption{Sample of \textsc{Bool8} data.}\label{tbl:bool8data}
\end{table*}

\begin{table*}[tb]
\begin{center}
\begin{tabular}{ccc}
    \multicolumn{3}{c}{\textsc{Poly8}}\\ \toprule
  $-a-c$ & $c^2$ & $b^2c^2$\\ \midrule
  $(b-a)-(c+b)$ & $(c\cdot c)+(b-b)$ & $(b\cdot b)\cdot(c\cdot c)$\\
  $b-(c+(b+a))$ & $((c\cdot c)-c)+c$ & $c\cdot (c\cdot (b\cdot b))$\\
  $a-((a+a)+c)$ & $((b+c)-b )\cdot c$ & $(c\cdot b) \cdot(b\cdot c)$\\
  $(a-(a+a))-c$ & $c\cdot (c-(a-a))$ & $((c\cdot b)\cdot c)\cdot b$\\
  $( b - b ) - ( a + c )$ & $c \cdot c$ & $((c \cdot c) \cdot b ) \cdot b$\\
\toprule
    	$c$ &  $b\cdot c$ & $b-c$\\ \midrule
	$c - ( ( c - c ) \cdot a )$ & $( c - ( b - b ) ) \cdot b$ & $( a - ( a + c ) ) + b$\\
	$c - ( ( a - a ) \cdot c )$ & $( b - ( c - c ) ) \cdot c$ & $( a - c ) - ( a - b )$\\
	$( ( a - a ) \cdot b ) + c$ & $( b - b ) + ( b \cdot c )$ & $( b - ( c + c ) ) + c$\\
	$( c + a ) - a$ & $c \cdot ( ( b - c ) + c )$ & $( b - ( c - a ) ) - a$\\
	$( a \cdot ( c - c ) ) + c$ & $( b \cdot c ) + ( c - c )$ & $b - ( ( a - a ) + c )$\\
  \bottomrule
\end{tabular}
\caption{Sample of \textsc{Poly8} data.}\label{tbl:poly8data}
\end{center}
\end{table*}

\section{Detailed Evaluation}
\label{appendix:detailedevaluation}
\begin{figure*}[tb]\centering
  \begin{subfigure}[b]{\textwidth}
    \includegraphics[width=\textwidth]{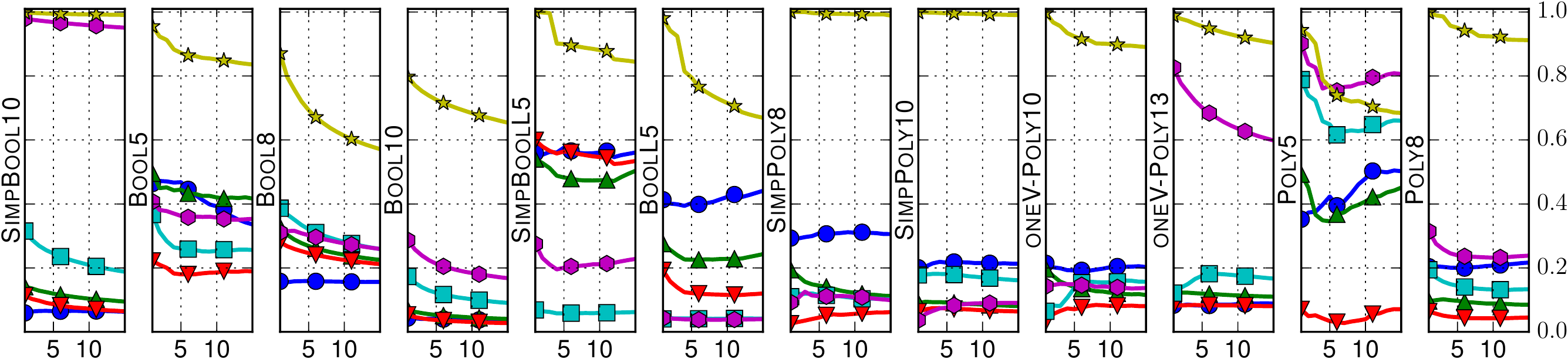}
    \caption{\seentest evaluation using model trained on
    the respective training set.}\label{fig:knnEvalDetailed:self:test}
  \end{subfigure}
  \begin{subfigure}[b]{\textwidth}
    \includegraphics[width=\textwidth]{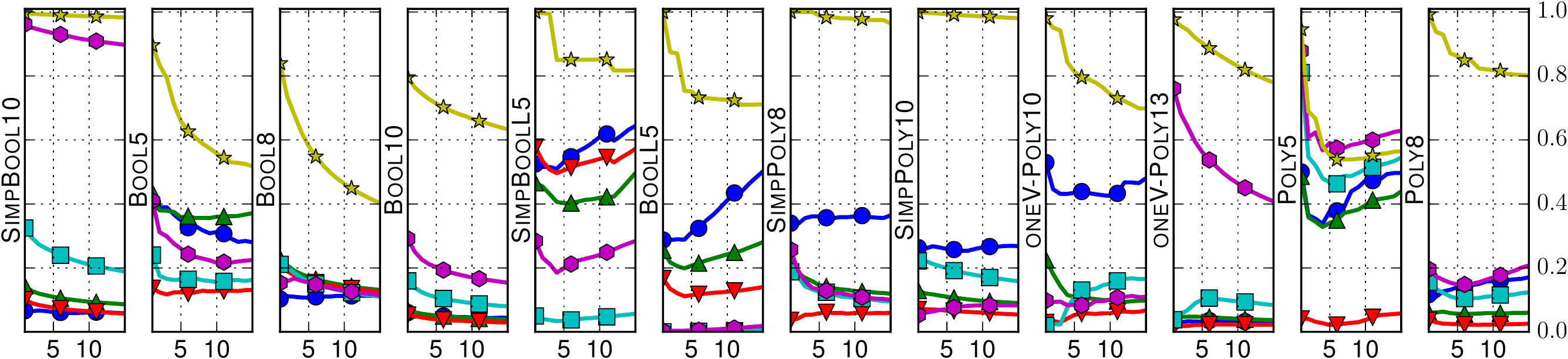}
    \caption{\unseentest evaluation using model trained on
    the respective training set.}\label{fig:knnEvalDetailed:self:neweq}
  \end{subfigure}
  \caption{Evaluation of $\knnscore{x}$ ($y$ axis) for $x=1,\dots,15.$ on the
  respective \seentest and \unseentest where each model has been trained on. The markers
  are shown every five ticks of the $x$-axis to make the graph more clear. \recNN
  refers to the model of \citet{socher2012semantic}.}\label{fig:knnEvalDetailed}
\end{figure*}

\begin{figure*}[tb]\centering
  \begin{subfigure}[b]{\textwidth}
    \includegraphics[width=\textwidth]{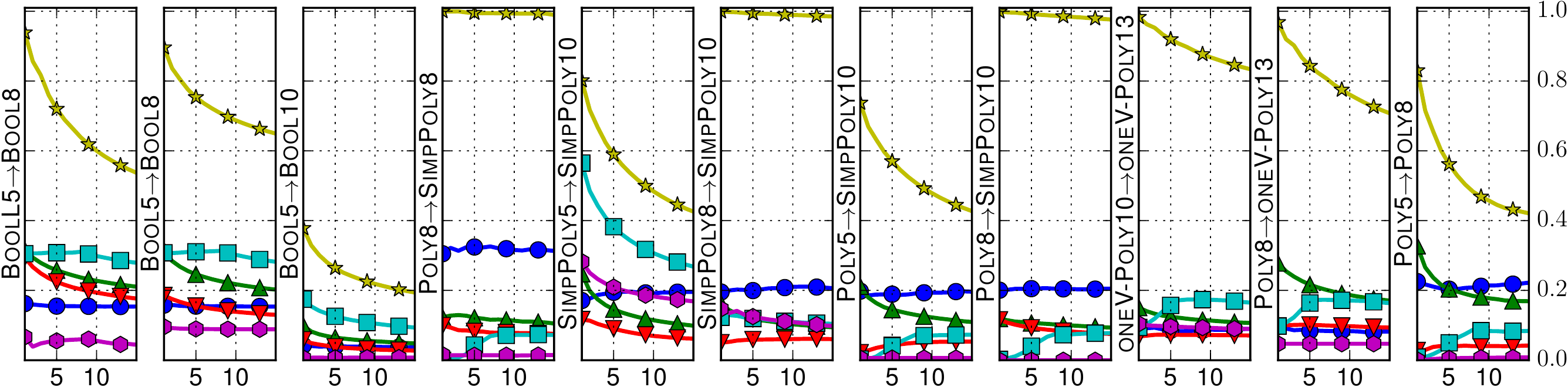}
    \caption{\seentest evaluation using model trained on simpler datasets.
    Caption is ``model trained on''$\rightarrow$``Test dataset''.}\label{fig:knnEvalDetailed:transfer:test}
  \end{subfigure}
  \begin{subfigure}[b]{\textwidth}
    \includegraphics[width=\textwidth]{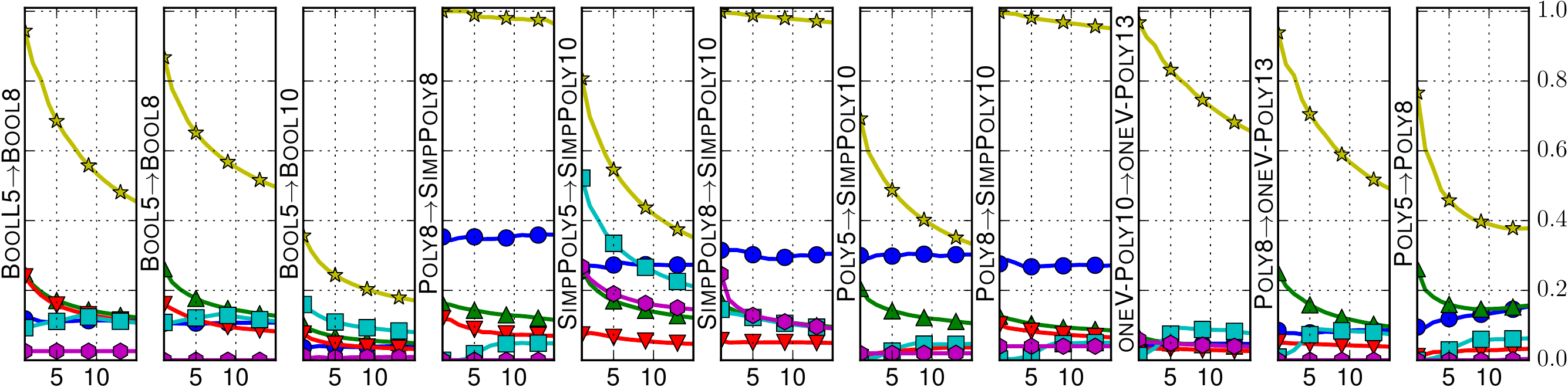}
    \caption{Evaluation of compositionality. \unseentest evaluation using model trained on simpler datasets.
    Caption is ``model trained on''$\rightarrow$``Test dataset''.}\label{fig:knnEvalDetailed:transfer:neweq}
  \end{subfigure}
  \includegraphics[width=\textwidth]{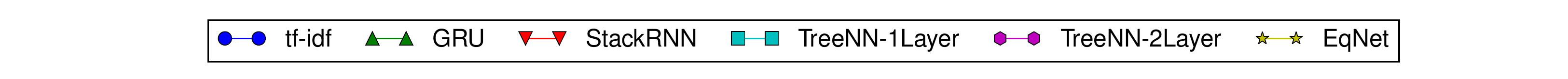}
  \caption{Evaluation of compositionality. Evaluation of $\knnscore{x}$ ($y$ axis) for $x=1,\dots,15.$ The markers
  are shown every five ticks of the $x$-axis to make the graph more clear. \recNN
  refers to the model of \citet{socher2012semantic}.}\label{fig:knnEvalDetailed:transfer}
\end{figure*}

\begin{figure*}[tb]
  \begin{subfigure}[b]{.4\textwidth}\centering
    \includegraphics[width=.8\textwidth]{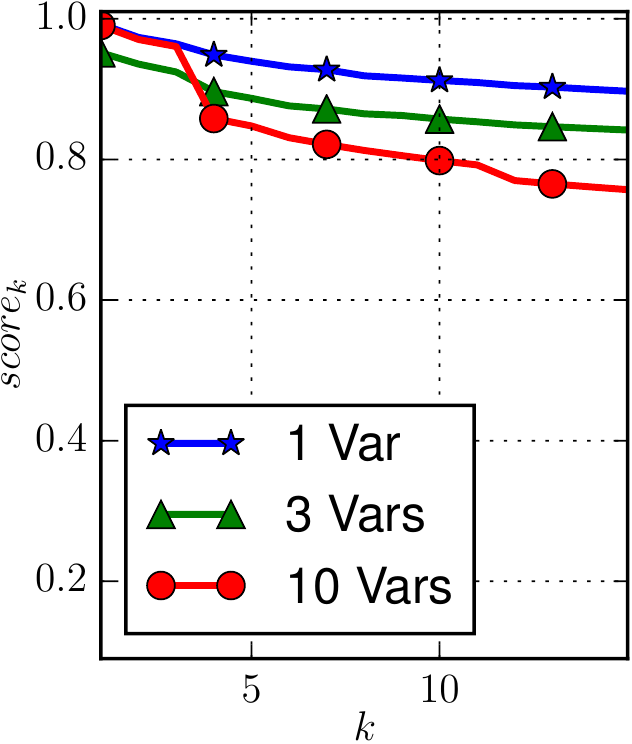}
    \caption{Performance \vs Number of Variables}\label{fig:knnEvalPerVar}
  \end{subfigure}\hfill
  \begin{subfigure}[b]{.4\textwidth}\centering
    \includegraphics[width=.8\textwidth]{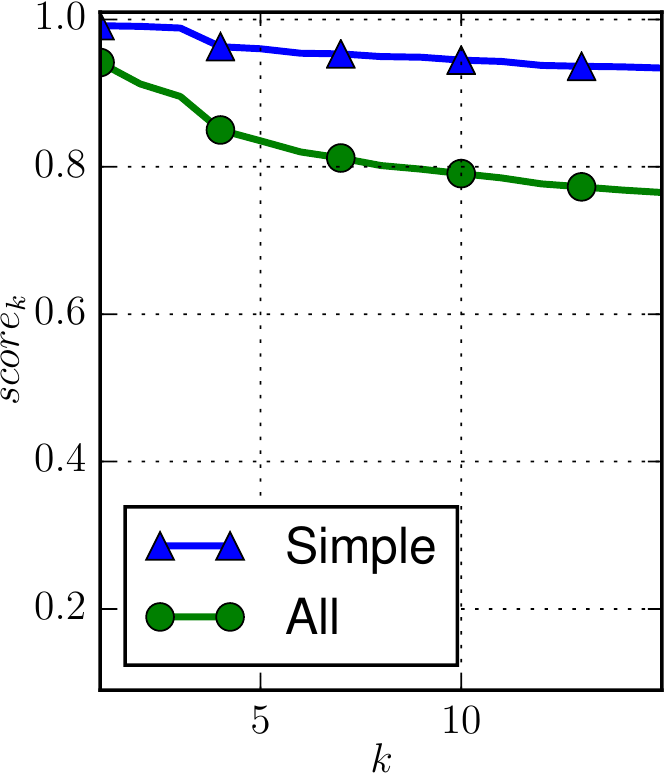}
    \caption{Performance \vs Operator Complexity}\label{fig:knnEvalPerOpType}
  \end{subfigure}

  \caption{\eqnet performance on \seentest for various dataset characteristics}\label{fig:perfForCharacteristic}
\end{figure*}

\autoref{fig:knnEvalDetailed} presents the detailed evaluation for our $k$-NN metric
for each dataset. \autoref{fig:knnEvalDetailed:transfer} shows the detailed
evaluation when using models trained on simpler datasets but tested on more
complex ones, essentially evaluating the learned compositionality of the models.
\autoref{fig:perfForCharacteristic} show how the performance varies across
the datasets based on their characteristics. As expected as the number
of variables increase, the performance worsens (\autoref{fig:knnEvalPerVar})
and expressions with more complex operators tend to have worse performance
(\autoref{fig:knnEvalPerOpType}). The results for \unseentest look
very similar and are not plotted here.

\begin{figure*}[tb]
  \begin{subfigure}[b]{.48\textwidth}\centering
    \includegraphics[width=\textwidth]{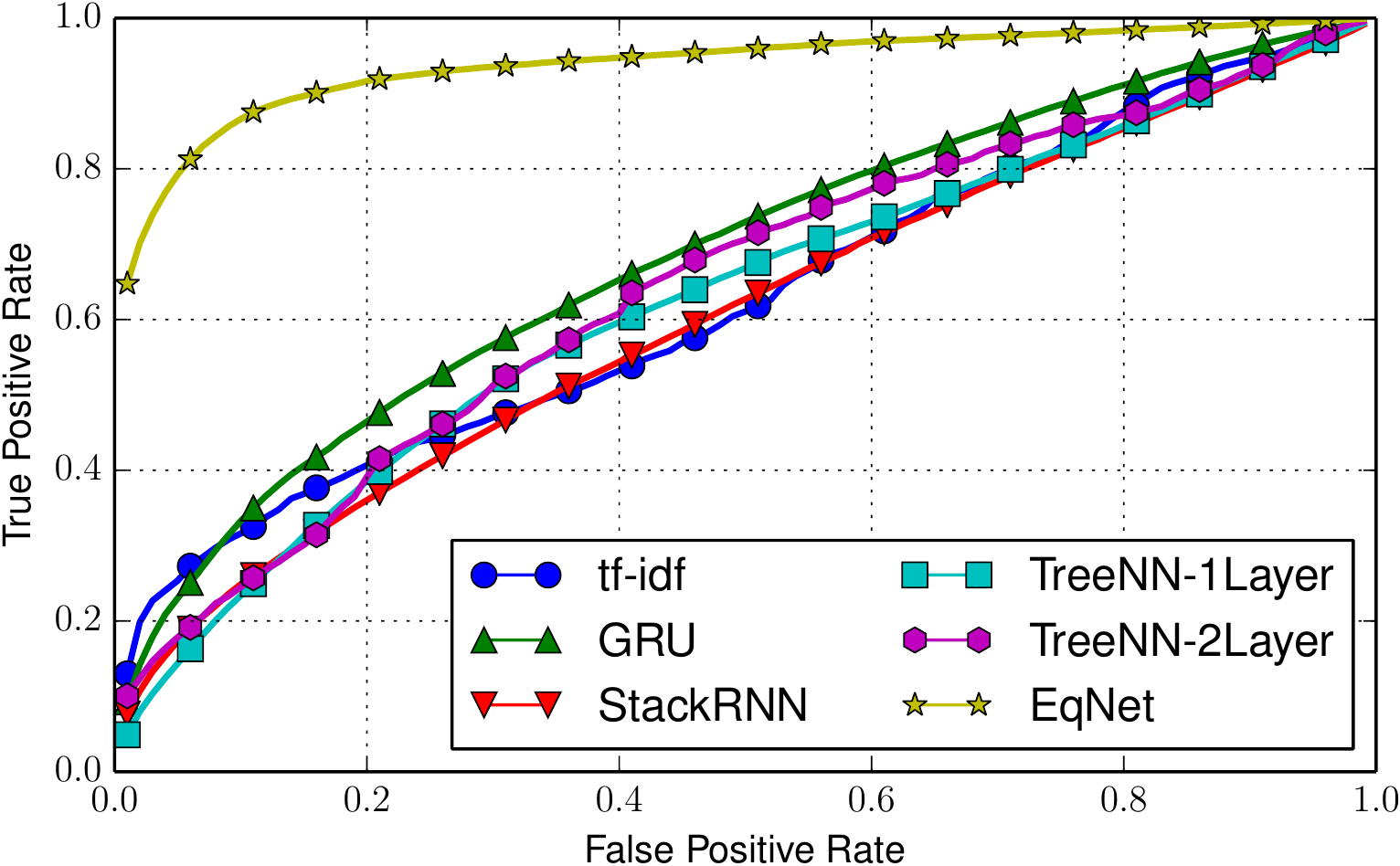}
    \caption{\seentest}\label{fig:rocSeen}
  \end{subfigure}\hfill
  \begin{subfigure}[b]{.48\textwidth}\centering
    \includegraphics[width=\textwidth]{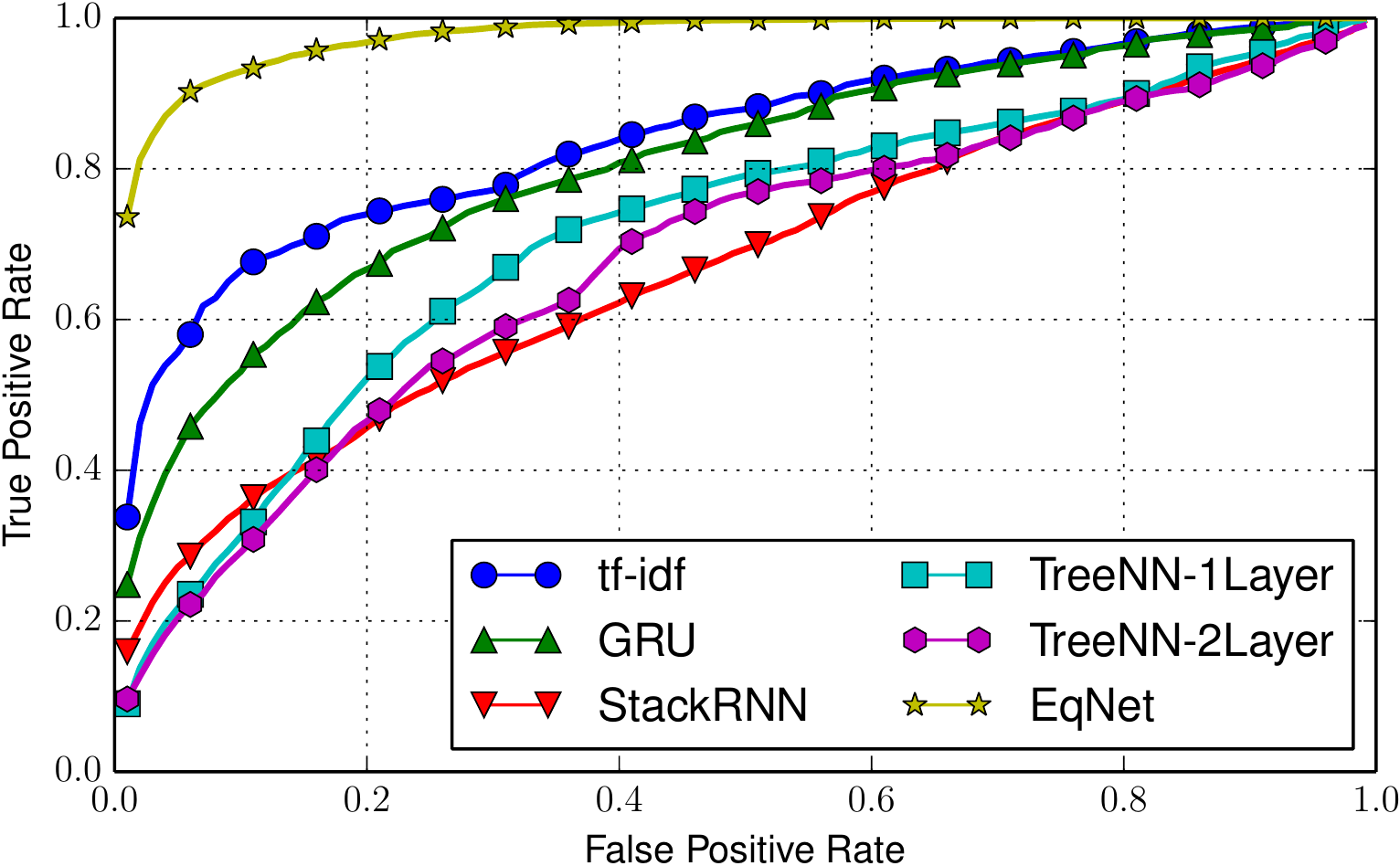}
    \caption{\unseentest}\label{fig:rocUnseen}
  \end{subfigure}
  \caption{Receiver operating characteristic (ROC) curves averaged across datasets.}\label{fig:rocCurves}
\end{figure*}

\section{Model Hyperparameters}
\label{appendix:hyperparameters}
The optimized hyperparameters are detailed in \autoref{tbl:hyperparams}. All hyperparameters were
optimized using the Spearmint \citep{snoek2012practical} Bayesian optimization
package. The same range of values was used for all common model hyperparameters.

\begin{table*}[tb] \centering
\caption{Hyperparameters used in this work.}\label{tbl:hyperparams}
\begin{tabular}{lp{12cm}} \toprule
    Model & Hyperparameters \\ \midrule
    \eqnet & learning rate $10^{-2.1}$, rmsprop $\rho=0.88$, momentum $0.88$, minibatch size 900, representation size $D=64$,
          autoencoder size $M=8$, autoencoder noise $\kappa=0.61$, gradient clipping $1.82$, initial parameter standard deviation $10^{-2.05}$,
          dropout rate $.11$, hidden layer size $8$, $\nu=4$, curriculum initial tree size $6.96$, curriculum step per epoch $2.72$,
          objective margin $m=0.5$ \\
    1-layer-\recNN & learning rate $10^{-3.5}$, rmsprop $\rho=0.6$, momentum $0.01$, minibatch size 650, representation size $D=64$,
          gradient clipping $3.6$,  initial parameter standard deviation $10^{-1.28}$, dropout $0.0$,
           curriculum initial tree size $2.8$, curriculum step per epoch $2.4$, objective margin $m=2.41$ \\
    2-layer-\recNN & learning rate $10^{-3.5}$, rmsprop $\rho=0.9$, momentum $0.95$, minibatch size 1000, representation size $D=64$,
          gradient clipping $5$,  initial parameter standard deviation $10^{-4}$, dropout $0.0$, hidden layer size $16$,
           curriculum initial tree size $6.5$, curriculum step per epoch $2.25$, objective margin $m=0.62$ \\
    GRU & learning rate $10^{-2.31}$, rmsprop $\rho=0.90$, momentum $0.66$, minibatch size 100, representation size $D=64$,
          gradient clipping $0.87$, token embedding size $128$, initial parameter standard deviation $10^{-1}$,
          dropout rate $0.26$\\
    StackRNN & learning rate $10^{-2.9}$, rmsprop $\rho=0.99$, momentum $0.85$, minibatch size 500, representation size $D=64$,
          gradient clipping $0.70$, token embedding size $64$, RNN parameter weights initialization standard deviation $10^{-4}$,
          embedding weight initialization standard deviation $10^{-3}$, dropout $0.0$, stack count $40$\\
  \bottomrule
\end{tabular}
\end{table*}

\end{document}